\documentclass{bmvc2k}
\usepackage{hyperref}
\usepackage{geometry}
\usepackage{graphics}
\usepackage{amsthm}
\usepackage{mathrsfs}
\usepackage{palatino,latexsym,graphicx,color}
\usepackage{times, amsmath,amssymb,booktabs}
\usepackage{multirow}
\usepackage{threeparttable}

\title{RGB-T Multi-Modal Crowd Counting \\Based on Transformer}

\addauthor{Zhengyi Liu}{liuzywen@ahu.edu.cn}{1}
\addauthor{Wei Wu}{2640947588@qq.com}{1}
\addauthor{Yacheng Tan}{1084043983@qq.com}{1}
\addauthor{Guanghui Zhang}{2532950974@qq.com}{1}
\addinstitution{
 School of Computer Science and Technology\\
 Anhui University\\
 Hefei, China
}

\runninghead{Liu \etal}{RGB-T Multi-Modal Crowd Counting}

\def\etal{\emph{et al}\bmvaOneDot}

\begin{document}

\maketitle

\begin{abstract}
Crowd counting aims to estimate the number of persons in a scene.
Most state-of-the-art crowd counting methods based on color images can't work well in poor illumination conditions due to invisible objects. With the widespread use of infrared cameras, crowd counting based on color and thermal images is studied. Existing methods only achieve multi-modal fusion without count objective constraint. To better excavate multi-modal information, we use count-guided multi-modal fusion and modal-guided count enhancement to achieve the impressive performance. The proposed count-guided multi-modal fusion module utilizes a multi-scale token transformer to interact two-modal information under the guidance of count information and perceive different scales from the token perspective. The proposed modal-guided count enhancement module employs multi-scale deformable transformer decoder structure to enhance one modality feature and count information by the other modality. Experiment in public RGBT-CC dataset shows that our method refreshes the state-of-the-art results. \href{https://github.com/liuzywen/RGBTCC}{https://github.com/liuzywen/RGBTCC}
\end{abstract}

%-------------------------------------------------------------------------
\section{Introduction}
Crowd counting can predict the distribution of crowd and estimate the number of persons in unconstraint scenes.
It is widely studied by the academia and industrial communities since the number of persons is an important indicator of incident monitoring\cite{usman2021abnormal}, traffic control\cite{liu2020dynamic}, and infectious disease
prevention\cite{velavan2020covid}.
The existing crowd counting methods have achieved tremendous improvement due to the introduce of convolutional neural networks \cite{gao2020cnn,fan2022survey} and transformer\cite{sun2021boosting,wei2021scene}.

\begin{figure*}[!htp]
  \centering
  \includegraphics[width=1\linewidth]{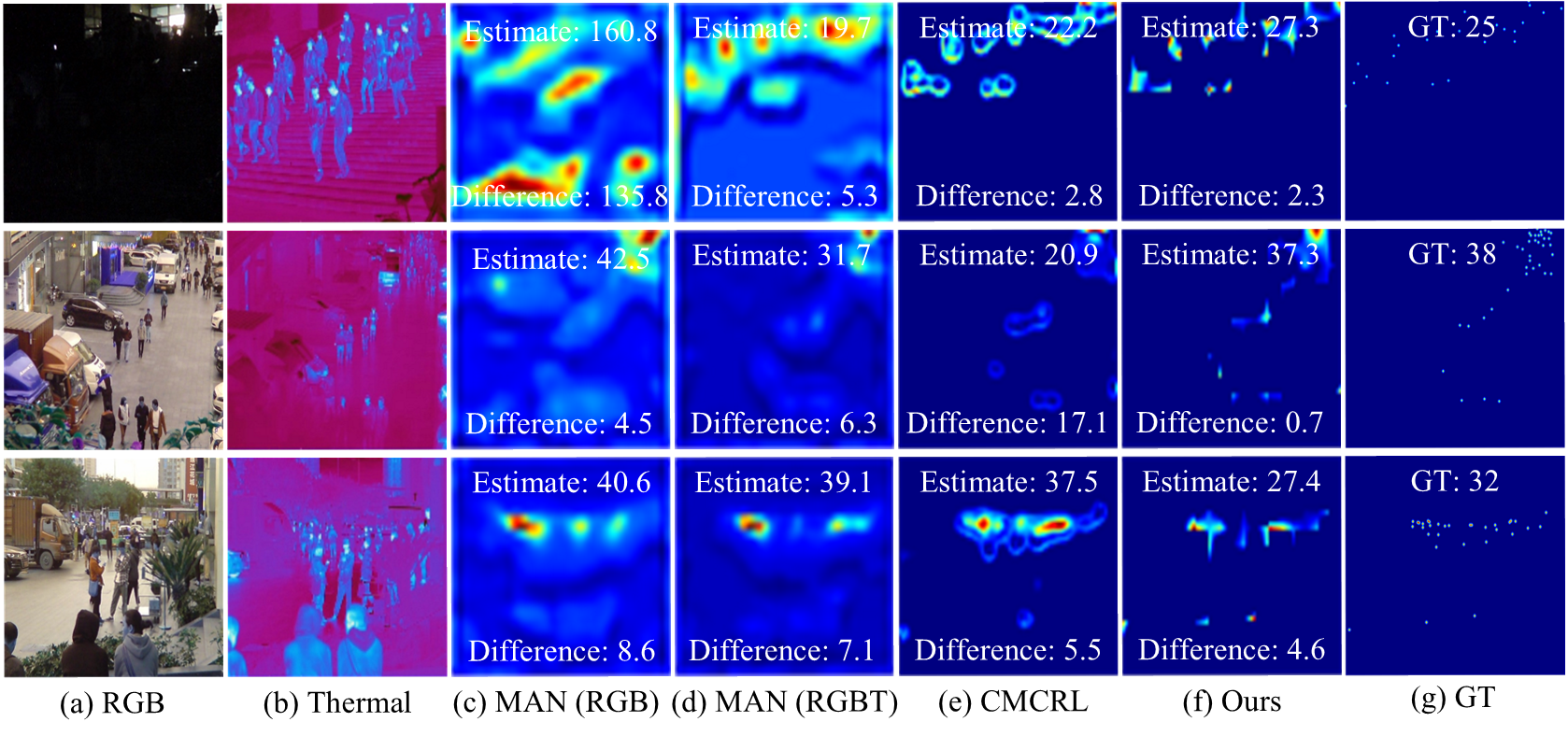}
  \caption{Three examples to show the performances of different methods in poor light condition, thermal disturbance, and large-scale variation, respectively.
  ``Estimate" means predicted counts. ``Difference" means the counting difference from the ground truth. (a) RGB image (b) the paired thermal image (c) MAN\cite{lin2022boosting} result based on RGB image (d) MAN\cite{lin2022boosting} result based on RGB-T image (e) CMCRL\cite{liu2021cross} result (f) our result (g) ground truth.
  \label{fig:examples}}
\end{figure*}

However, when light is insufficient, the performance of crowd counting is unsatisfying, as shown in the first line of Fig.\ref{fig:examples}. The thermal image can percept the  temperature  of objects to recognize the persons. Therefore, RGB-Thermal (RGB-T) crowd counting by introducing the thermal modality has attracted a lot of attentions.

In RGB-T crowd counting task, a most important challenge is the multi-modal fusion problem. The color modality is good at perceiving the shape and texture of persons, but it is also interfered  by the cluttered background.
The thermal modality is skilled in recognizing persons which have temperature from scattered environments, but it also highlights the other heating objects, as shown in the second line of Fig.\ref{fig:examples} where the cars in the right are highlighted.
Existing RGB-T crowd counting methods fuse complementary multi-modal features by Information Aggregation and
Distribution Module (IADM) \cite{liu2021cross}, Information Improvement Module (IIM) \cite{tang2022tafnet}, and Mutual Attention Transformer (MAT) \cite{wu2022multimodal}. However, these multi-modal interactions are lack of a constraint. If we add a counting constraint on multi-modal fusion process, two-modal fusion has a clear goal.
Therefore, we use a transformer structure to fuse two-modal information and design a learnable count token to participant the two-modal fusion. It makes the color and thermal modality interact under the guidance of a common count token.

In RGB-T crowd counting task, the other challenge is large-scale variation which is also the common issue in crowd counting, as shown in the third line of Fig.\ref{fig:examples} where persons that are far from the camera appear much smaller than those close to it. %The size of persons are different due to their distances from the camera.
Existing methods use multi-column structure \cite{zhang2016single,babu2017switching,liu2019context,yu2022frequency}, dilated convolution \cite{li2018csrnet,bai2020adaptive,dai2021dense}, high-resolution representation \cite{sajid2021audio}, and attention mechanism \cite{lin2022boosting} to enlarge the receptive fields. Under the transformer framework, we propose a multi-scale token transformer to perceive persons with different scales. The tokens are merged to form token sequences with different lengths and then fed into some parallel transformers. After the enhancement of transformers, the receptive fields of features will be diversified.

To further improve the accuracy of crowd counting, we use a modality to guide the learning of the other modality and count token. A multi-scale deformable transformer is adopted to decode a modality and count token by the other modality. As a result, the count ability of the feature is enhanced.

In summary, the main contributions are summarized as follows:
\begin{itemize}
  \item
  An RGB-T multi-modal crowd counting model is proposed based on the transformer. Multi-head self-attention is used to achieve the count-guided multi-modal fusion. Multi-head cross-attention is adopted to achieve the modal-guided count enhancement.
  \item
  A count-guided multi-modal fusion transformer is proposed to solve the fusion problem. Under the guidance of count global information, color and thermal modalities are well combined and aligned.
  \item
  A multi-scale token transformer is proposed to solve the large-scale variation problem. Three-scale token sequences are parallel handled to achieve multi-scale concept.
  \item
  The ablation experiments verify the effectiveness of modules, multi-scale design, and count guidance. The comparison experiments show the significant improvement over existing RGB-T crowd counting methods.
\end{itemize}

\section{Related work}
\subsection{Crowd counting}
Crowd counting can be achieved by detection \cite{hoai2014talking,stewart2016end,idrees2015detecting,lian2021locating} or density map estimation \cite{wang2022crowd,jiang2020density,wang2022stnet,chen2022ssr}.
Since the latter can solve high overlap and occlusion problem, it shows better performance than the former.

%%%%%%%%%%%%%%%%%
The large-scale variation generated by the wide viewing angle of cameras and 2D perspective projection is a major challenge in crowd counting.
The persons which are close to the camera are large, while the persons which are far from the camera are small.
Multi-scale architecture\cite{zhang2016single,babu2017switching,li2018csrnet,yuan2020crowd,bai2020adaptive,dai2021dense} and perspective information\cite{shi2019revisiting,yan2019perspective,yang2020reverse,gao2019pcc,yang2020embedding} are two main solutions.
%CFANet\cite{rong2021coarse} filter the backbone noise.
Recently, to deal with the scale changes, some attention based methods are proposed.
%CDADNet\cite{zhu2021cdadnet} suppresses the background by adding attention map to density expansion network.
MAN\cite{lin2022boosting} improves global attention in the transformer by adding region attention.
HANet\cite{wang2022hybrid} introduces scale context in the parallel spatial attention and channel attention.
%AEDNet\cite{liu2022attentive} introduces attention into an encoder-decoder network.
%AMS-Net\cite{zhang2021crowd} refines density map by adding an attention subnetwork.
%SCLNet\cite{wang2020sclnet} incorporates a parallel spatial and channel attention in the decoder.
%There are also some multi-task methods.
%PDANet\cite{amirgholipour2021pdanet} introduces classification task to guide density map estimation.
%ASANet\cite{chen2021adversarial} incorporates object detection task in the adversarial learning framework.
%MMCNN\cite{yang2018counting} integrates density level estimation and background and foreground mask estimation.
%In addition, %RRP\cite{chen2020relevant} leverages the graph convolutional network to capture region dependency with different densities.
%SASNet\cite{song2021choose} obtains the scale information by patch-wise feature level selection.
%%DSNet\cite{dai2021dense} uses dense dilated convolution block to model the continuously varied scale.
%RANet\cite{chen2022region} embeds the relevance between  input image and priority map into  input image to understand scale variation.
%STNet \cite{wang2022stnet} designs a tree structure to organize different atrous convolutions for the broader  range of receptive fields.

In the paper, we solve the large-scale variation problem by multi-scale transformer based on tokens. The original token sequence is merged into a middle-scale token sequence and a large-scale token sequence, respectively. Then the three are parallel handled by three multi-head self-attention structures. Finally, three branches are concatenated and combined. The multi-scale concept ensures abundant receptive fields which benefits the crowd counting task.

\subsection{Transformer based crowd counting}
Previous works utilize the convolution neural network as the backbone and regress density map to predict the crowd count.
The advent of transformer  has pushed the crowd counting model forward.
%Boosting Crowd Counting with Transformers
BCCTrans \cite{sun2021boosting} introduces a global context learnable token to guide the counting.
%(SAANet)Scene-Adaptive Attention Network for Crowd Counting
SAANet \cite{wei2021scene} designs a deformer backbone to extract the features, aggregates multi-level features by a deformable transformer encoder, and introduces a count query in a transformer decoder to re-calibrates the multi-level feature maps.
%%Congested Crowd Instance Localization with Dilated Convolutional Swin Transformer
DCSwinTrans\cite{gao2021congested} enhances the large-range contextual information by a dilated Swin Transformer backbone, and equips with a feature pyramid networks decoder to achieve crowd instant localization.
CrowdFormer \cite{yangcrowdformer} models the human top-down
visual perception mechanism by an overlap patching transformer block.
%fully and weakly supervised
%CCTrans: Simplifying and Improving Crowd Counting with Transformer
CCTrans \cite{tian2021cctrans} adopts a pyramid transformer and a multi-scale regression head to achieve both fully-supervised and weakly-supervised crowd counting task.
%%%%%%%%%%%% weakly
In addition, in weakly-supervised crowd counting, there are some other transformer based methods.
TransCrowd \cite{liang2022transcrowd} uses a learnable counting token or global average pooling on high-layer semantic tokens to represent the crowd numbers. It constructs a weakly supervised model from sequence-to-count perspective.
%Reinforcing Local Feature Representation for Weakly-Supervised Dense Crowd Counting (good)
SFSL \cite{chen2022reinforcing} introduces a learnable unbiased feature estimation of persons and utilizes the feature similarity for the regression of crowd numbers to solve the lack of local supervision.
%CrowdMLP
CrowdMLP \cite{wang2022crowdmlp} proposes a multi-granularity multilayer perceptron (MLP) regressor to enlarge receptive fields and a split-counting to decouple spatial constraints.
%Joint CNN and Transformer Network via weakly supervised Learning for efficient crowd counting
JCTNet \cite{wang2022joint} introduces transformer structure upon the high-layer feature of convolutional neural network and regresses the count.

In the paper, we use transformer encoder structure to achieve count-guided multi-modal fusion, and use transformer decoder structure to perform modal-guided count enhancement.

\subsection{RGB-T crowd counting}
Although the crowd counting methods have achieved
many significant improvements, they rely on optical information and often perform poorly when the light is insufficient.
To solve this problem, RGB-T crowd counting has been getting a lot of attentions.
On one hand, thermal image can recognize pedestrians in poor illumination conditions. On the other hand, thermal image can reduce wrong recognition about some human-shaped objects.
Meanwhile, RGB image can suppress interference in thermal images. For example, heating walls and lamps that are highlighted in thermal images can be filtered from color perspective.
Therefore, RGB and thermal images need to be simultaneously explored.

%More recently, RGB-T crowd counting is becoming an active research by the open of RGBT-CC \cite{liu2021cross} crowd counting datasets.
%MMCCN \cite{peng2020rgb} introduced a multi-modal crowd counting network that uses generative adversarial network to align two-modal features and finally adaptively fuses two-modal results.
%MMCCN \cite{peng2020rgb} aligns two-modal features by generative adversarial network and obtains the final result by adaptive fusion.
%Cross-Modal Collaborative Representation Learning and a Large-Scale RGBT Benchmark for Crowd Counting
%CMCRL\cite{liu2021cross} incorporates  multiple modality-specific branches, a modality-shared branch to fully capture the complementarities among different modalities for RGB-T crowd counting.
CMCRL \cite{liu2021cross} introduces  a two-stream framework  that first aggregates two features and second propagates the common information to further refine each feature.
TAFNet \cite{tang2022tafnet} uses a three-stream network to learn the RGB feature, the thermal feature, and the concatenated RGB-T feature for crowd counting. The proposed Information Improvement Module (IIM) is used to fuse the modal-specific and combination features.
Mutual Attention Transformer (MAT) \cite{wu2022multimodal} uses cross-modal mutual attention to build long-range dependencies and enhance semantic features in crowd counting task.
DEFNet \cite{zhou2022defnet} uses multi-modal fusion, receptive field enhancement, and multi-layer fusion to highlight the crowd position and suppress the background noise.

In these works, the fusion of the RGB and thermal images are  short of count objective constraint. We design a learnable count token to guide multi-modal fusion. %Two-modal features are combined under the guidance of a common objective. Thus, two-modal fusion is with a global goal.

\section{Proposed Method}
\begin{figure*}[!htp]
\center
  \includegraphics[width=0.9\textwidth]{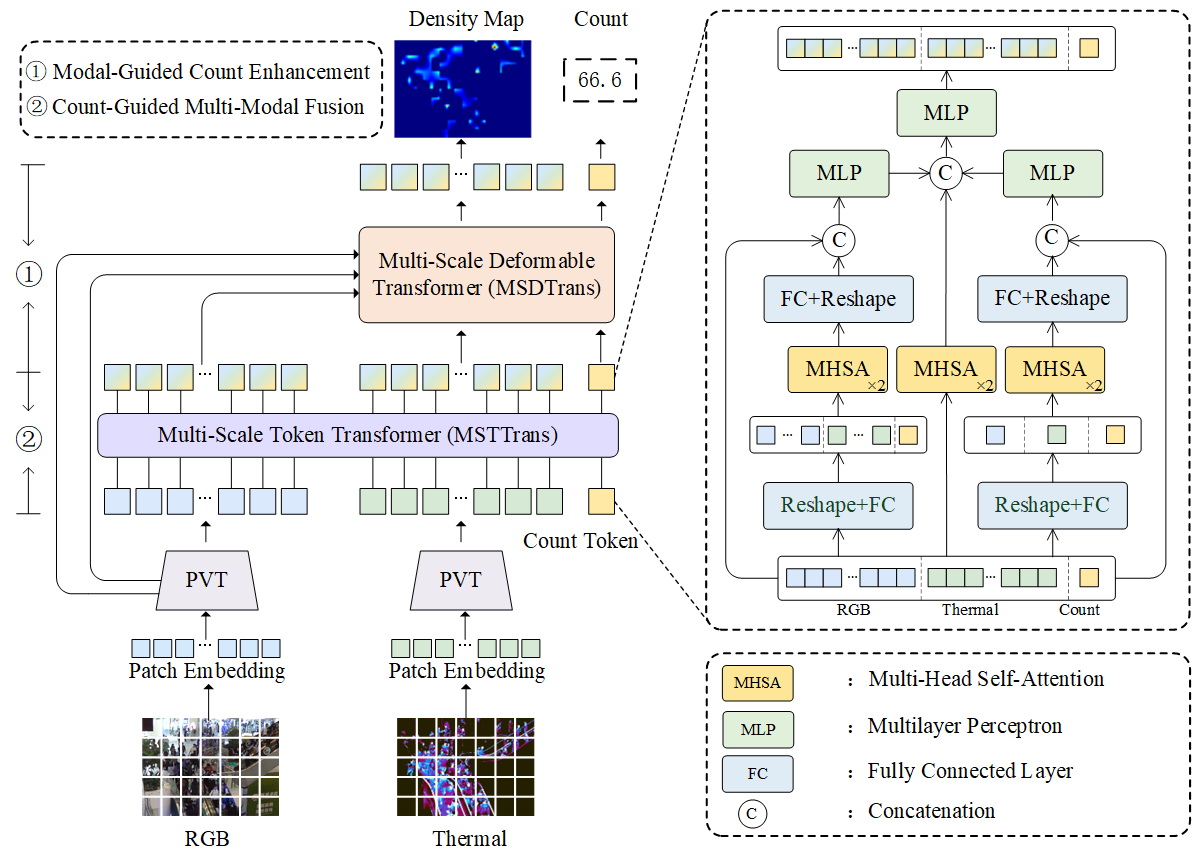}
  \caption{Our proposed RGB-T multi-modal crowd counting model based on transformer.}
  \label{fig:main}
\end{figure*}

We propose an RGB-T multi-modal crowd counting method which includes a count-guide multi-modal fusion, a modal-guide count enhancement, and a regression head, as shown in Fig.\ref{fig:main}. To solve multi-modal fusion problem, we introduce a count guidance. Moreover, to perceive the large-scale variation, we propose a multi-scale token concept. Combining both, multi-modal features are well fused towards a global objective. Furthermore, counting information is further enhanced from one modality under the guidance of the other modality.
\subsection{Count guided multi-modal fusion}
Given a paired RGB-T image  $I=\{I_r,I_t\}$, we use two PVT encoders \cite{wang2021pyramid} as the feature extractors to capture hierarchical features.
\begin{equation}
\begin{aligned}
F_r=\mathcal {E}_{PVT}(I_r)\\
F_t=\mathcal {E}_{PVT}(I_t)\\
\end{aligned}
\end{equation}
where $\mathcal {E}_{PVT}$ denotes a PVT encoder, $F_r=\{F_r^i\}_{i=1}^{4}$ and $F_t=\{F_t^i\}_{i=1}^{4}$ represent color features and thermal features, respectively, $i$ is the feature layer number.

The high-layer features contain more semantic information, which are suitable to obtain the global counting cues.
Besides, color feature and thermal feature have each advantage in representing the crowd.
Therefore, we use the high-level tokens from color modality and thermal modality to excavate the number of crowd.
To fully align two-modal data and generate a consistent result, a learnable count token is designed to guide the two-modal fusion.
Specifically, as is illustrated in Fig.\ref{fig:main}, high-layer semantic features $F_r^4$ and $F_t^4$ are generated from color encoder and thermal encoder, respectively.
They represent unaligned multi-modal semantic concept.
We design a learnable count token $F_{count}$ which implies the coarse number of crowd. The three are concatenated along the token direction, and then fed into a Multi-Scale Token Transformer ($\textit{MSTTrans}$) which spreads information among color, thermal, and crowd count by the multi-head self-attention.

$\textit{MSTTrans}$ is proposed to solve large-scale variations.
Inspired by multi-scale design in  atrous spatial pyramid pooling (ASPP) \cite{chen2017deeplab},
$\textit{MSTTrans}$ achieves multi-scale transformer based on tokens.
At first, we concatenate high-layer color feature, high-layer thermal feature, and the learnable count token to form an initial token sequence.
Then, we merge the initial token sequence to generate a middle-scale token sequence. The middle-scale token sequence has the larger receptive fields than original token sequence. Besides, we merge the initial  token sequence to generate a large-scale token sequence, where a modality is represented by a token.
According to above merge strategy, three parallel branches which all include color modality, thermal modality, and count token are constructed.
They are fed into three multi-head self-attention modules for in-depth fusion.

Specifically,  as is illustrated in the right of Fig. \ref{fig:main}, suppose the high-layer semantic feature $F_r^4  \in \mathbb{R}^{N^2\times C}$ and $F_t^4  \in \mathbb{R}^{N^2\times C}$, where $N^2$ and $C$ represent the number of tokens and channels, respectively.
The two-modal features and the learnable count token are concatenated to generate the initial token sequence $f_1\in \mathbb{R}^{(2N^2+1)\times C}$.
\begin{equation}
\begin{aligned}
f_1=[F_r^4,F_t^4,F_{count}]
\end{aligned}
\end{equation}
where $[\cdot]$ denotes concatenation operation along token direction.

Then, the two-modal features are merged to $N$ groups and each group generates $N$ middle-scale tokens. All the middle-scale tokens and the learnable count token are concatenated to generate the middle-scale token sequence $f_2\in \mathbb{R}^{(2N+1)\times C}$.
\begin{equation}
\begin{aligned}
f_2=[\textit{merge}_{N^2\rightarrow N}(F_r^4),\textit{merge}_{N^2\rightarrow N}(F_t^4),F_{count}]
\end{aligned}
\end{equation}
where $\textit{merge}_{a\rightarrow b}$ denotes the aggregation operation from $a$ tokens to $b$ tokens which applies a reshape operation and a fully connected layer.

Meanwhile, the two-modal features are merged to two groups and each group generates a large-scale token.
The large-scale tokens and the learnable count token are concatenated to generate the large-scale token sequence $f_3\in \mathbb{R}^{(2+1)\times C}$.
There are a color token, a thermal token, and a learnable count token. It ensures two-modal whole alignment under the guidance of count.
\begin{equation}
\begin{aligned}
f_3=[\textit{merge}_{N^2\rightarrow 1}(F_r^4),\textit{merge}_{N^2\rightarrow 1}(F_t^4),F_{count}]\\
\end{aligned}
\end{equation}

Three  token sequences with different scales are fed into three multi-head self-attention modules for multi-modal interaction.
\begin{equation}
\begin{aligned}
f_i^\prime=\textit{MHSA}(f_i)
\end{aligned}
\end{equation}
where $\textit{MHSA}$ represents two multi-head self-attention layers.

Since the lengths of middle-scale and large-scale token sequences are different from initial token sequences, we apply fully connection layer and reshape operation to restore token sequence length.
\begin{equation}
\begin{aligned}
g_i=\textit{Reshape}(\textit{FC}(f_i^\prime))
\end{aligned}
\end{equation}
where $i=2,3$ because only middle-scale and large-scale token sequences should be restored, $\textit{FC}$ is a fully-connected layer, and $\textit{Reshape}$ is a reshape operation to restore token length.

Further, to retain the original features in the middle-scale and large-scale branches, the concatenation and $\textit{MLP}$ operations are successively conducted.
\begin{equation}
\begin{aligned}
g_i^\prime=\textit{MLP}(\textit{Concat}(g_i,f_1))
\end{aligned}
\end{equation}
where $i=2,3$, \textit{Concat} is concatenation operation along channel direction, and $\textit{MLP}$ is a two-layer perceptron.

Last, three features are concatenated and  shrunk in channels.
\begin{equation}
\begin{aligned}
G=[G_r,G_t,G_{count}]=\textit{MLP}(\textit{Concat}(f_1^\prime,g_2^\prime,g_3^\prime))
\end{aligned}
\end{equation}
where $G$ has the same size  as the input $f_1$ of $\textit{MSTTrans}$ module, and consists of optimized color feature $G_r$, thermal feature $G_t$, and count feature $G_{count}$.

In $\textit{MSTTrans}$ module, the count token is responsible for incorporating the global information and perceiving the number of persons. Besides, it is used to guide the fusion of color feature and thermal feature.
Under the guidance of count token, color feature and thermal feature are in-depth interacted. Moreover, multi-scale token concept ensures the abundant receptive fields adaptive to recognizing the  persons with different sizes.

\subsection{Modal-guided counting enhancement}
The researches pointed out that the thermal image can provide strong support on density map estimation, especially in the dark background\cite{tang2022tafnet}.
In the paper, we use the thermal modality to predict the density map and count, and further use color modality to refine the prediction.

Therefore, after the previous count-guided multi-modal fusion, we design a modal-guided counting enhancement module which is responsible for generating the density map and final count from one modality under the guidance of the other modality. A multi-scale  deformable transformer ($\textit{MSDTrans}$) is employed to achieve the above objective.

Specifically, the thermal feature $G_t$ and the learnable count token $G_{count}$ are concatenated as query ($Q$), and the enhanced color feature $G_r$ and the encoded low-layer features $F_r^i(i=1,2,3)$ compose multi-scale color features which are regarded as key ($K$) and value ($V$). We use multi-scale deformable attention \cite{zhu2020deformable} to enhance $Q$ by $K$ and $V$. Last, it will output modal-guided enhanced feature $O_t$ and count token $O_{count}$.

\begin{equation}
\begin{aligned}
\left[O_t,O_{count}\right]=\textit{DeformAttn}(\left[G_t,G_{count}\right],\{G_r,F_r^3,F_r^2,F_r^1\})
\end{aligned}
\end{equation}
where $\textit{DeformAttn}(a,b)$ is the multi-scale deformable attention \cite{zhu2020deformable}, $a$ represents content feature, $b$ is multi-scale features.

\subsection{Regression head and loss function}
To obtain the density map, we use a simple regression head which consists of  two 3$\times$3 convolution layers and one 1$\times$1 convolution layer.
\begin{equation}
\begin{aligned}
D=\textit{RH}(O_t)
\end{aligned}
\end{equation}
where $\textit{RH}$ is the regression head.

The loss includes a loss about the density map and a loss about the learnable count token.
\begin{equation}
\begin{aligned}
\mathcal{L}=\mathcal{L}_{D}(D,D^\star)+\mathcal{L}_{C}(O_{count},C^\star)
\end{aligned}
\end{equation}
where $ \mathcal{L}_{D}$ adopts distribution matching loss  proposed in\cite{wang2020distribution}, which supervises the density map regression and count estimation, $\mathcal{L}_C$ adopts $L_1$ norm ($\Vert \cdot \Vert_1$) to supervise the count token. $D^\star$ and $C^\star$ represent the ground truth  of density map and count, respectively.

\section{Experiments}
\subsection{Datasets and evaluation metrics}
\textbf{Dataset.} The public RGBT-CC\cite{liu2021cross} dataset is adopted to evaluate our method. RGBT-CC consists of 1,030 training samples, 200  validation samples, and 800 testing ones.

\textbf{Evaluation Metrics.} The widely used Grid Average Mean Absolute Error (GAME)\cite{guerrero2015extremely} and Root Mean Square Error (RMSE) are used as evaluation metrics\cite{liu2021cross,tang2022tafnet}.
\begin{equation}
\begin{aligned}
GAME(l)=\frac{1}{N}\sum_{i=1}^N\sum_{j=1}^{4^l}\mid \hat{P}^j_i-P^j_i \mid
\end{aligned}
\end{equation}
where $\hat{P}^j_i$ represents the predicted value of the $j^{th}$ region of the $i^{th}$ image, $P^j_i$ indicates the ground truth corresponding to $\hat{P}^j_i$, $4^l$ means the number of the  divided non-overlapping regions of the image, and $N$ is the  number of paired images in testing dataset. $\textit{GAME}$ sums the counting errors in all the regions. %Note that when $l$ is 0, the $GAME(0)$ is equivalent to Mean Absolute Error (MAE).
\begin{equation}
\begin{aligned}
RMSE=\sqrt{\frac{1}{N}\sum_{i=1}^N(\hat{P}_i-P_i)^2}
\end{aligned}
\end{equation}
where $\hat{P}_i$ represents the predicted value of the $i^{th}$ image, $P_i$ indicates the ground truth corresponding to $\hat{P}_i$.
For both $\textit{RMSE}$ and $\textit{GAME}$, lower value means the better performance.
\subsection{Implementation details}
The implementation setting includes:
(1) GPU (NVIDIA RTX 3090); (2) input image size ($224\times224$); (3) train time (17 hours); (4) learning rate ($1e-5$); (5) weight decay ($1e-4$).
%Our model is trained on a NVIDIA RTX 3090 GPU. The input image size is $256\times256$.
%
%In the training phase, we random crop a batch of patches with the fixed size of from the original image. We use Adam optimizer with a learning rate of $1e-5$ and weight decay of $1e-4$ to train our model by minimizing the loss function\cite{wang2020distribution}. The max training epoch is set to 200 and batch size is set to 32. The train process takes about 8 hours.
\subsection{Comparison with state-of-the-art methods}
To make  quantitative comparisons, our  method is compared with recent prominent approaches, including CSRNet\cite{li2018csrnet}, BL\cite{ma2019bayesian}, DM-Count\cite{wang2020distribution}, P2PNet\cite{song2021rethinking}, MARUNet\cite{rong2021coarse},  MAN\cite{lin2022boosting}, CMCRL\cite{liu2021cross}, TAFNet \cite{tang2022tafnet}, MAT\cite{wu2022multimodal}, and DEFNet\cite{zhou2022defnet}  which are  shown in Table \ref{tab:comparison}.
The top of the table shows six single-modal crowd counting models which are retrained by the input fusion of RGB and thermal images.
The bottom of the table shows four RGB-T crowd counting models and ours.
From the observation, we can conclude our method performs the best among all the methods. It achieves about $8.4\%$, $7.8\%$, $5.7\%$, $4.1\%$, $10.9\%$ improvement over the second best result in GAME(0), GAME(1), GAME(2), GAME(3) and RMSE, respectively.
The great improvement profits from the multi-modal fusion under the guidance of count token and count enhancement of a modality under the guidance of the other modality.
%It reflects the powerful counting ability of our model in the RGB-T crowd counting.
\begin{table*}[!htp]
\caption{Comparison results of different methods on RGBT-CC benchmark dataset. \textbf{The top part: } some RGB crowd counting models are retrained by input fusion of color modality and thermal modality. \textbf{The bottom part: }some RGB-T crowd counting models. The best result is in bold.}
\centering
\resizebox{\linewidth}{!}
{
      \begin{tabular}{c|c|ccccc}
    \toprule
    Methods  &Source& GAME(0)$\downarrow$  &  GAME(1)$\downarrow$  &  GAME(2)$\downarrow$  &  GAME(3)$\downarrow$  &  RMSE$\downarrow$  \\
  \midrule
  CSRNet\cite{li2018csrnet} &CVPR2018 &  20.40  &  23.58  &  28.03  &  35.51  &  35.26  \\
  BL\cite{ma2019bayesian} &ICCV2019 &  18.70  &  22.55  &  26.83  &  34.62  &  32.67  \\
  DM-Count\cite{wang2020distribution}&NeurIPS2020&16.54&20.73&25.23&32.23&27.22\\
  P2PNet\cite{song2021rethinking}&ICCV2021&16.24&19.42&23.48&30.27&29.94\\
  MARUNet\cite{rong2021coarse}&WACV2021&17.39&20.54&23.69&27.36&30.84\\
  MAN\cite{lin2022boosting}&CVPR2022&17.16 & 21.78 & 28.74 & 41.59 & 33.84\\
  \midrule
  CMCRL\cite{liu2021cross} &CVPR2021 &  15.61  &  19.95  &  24.69  &  32.89  &  28.18  \\
  TAFNet\cite{tang2022tafnet} &ISCAS2022 &  12.38  &  16.98  &  21.86  &  30.19  &  22.45  \\

  % MAT(CSRNet)\cite{wu2022multimodal} &ICME2022 &  13.65 &18.03 &22.94& 28.65& 22.53  \\

  MAT\cite{wu2022multimodal} &ICME2022 &  12.35 &16.29 &20.81 &29.09 &22.53  \\
  DEFNet\cite{zhou2022defnet} &TITS2022 & 11.90 &16.08 &20.19 &27.27 &21.09  \\
  %MAFNet\cite{chen2022mafnet} &arxiv2022 & 10.90 &14.44 &18.36 &24.01 &20.82  \\

   Ours&  BMVC2022 &  \textbf{10.90}  &  \textbf{14.81}  &  \textbf{19.02}  &  \textbf{26.14}  &  \textbf{18.79} \\
  \bottomrule
  \hline
  \end{tabular}
}
\label{tab:comparison}
\end{table*}

\subsection{Ablation studies}
\subsubsection{Effectiveness analysis of the proposed modules}
To verify the effectiveness of the proposed modules, we conduct the ablation studies. Table \ref{tab:ModuleAblation} show the result.
At first, we construct a baseline model. It concatenates high-layer features of two PVT encoders and applies regression head to predict the density map and sum up. The baseline result is shown in the first line.
Then, we add count-guided multi-modal fusion module and modal-guided count enhancement module based on the baseline, respectively.
The result is shown in the second and the third lines, respectively.
Finally, we add all the modules. The result is shown in the fourth line.
By the observation, $\textit{MSTTrans}$ improves the performance from $\textit{GAME0}$ (11.62) to $\textit{GAME0}$ (10.91).
It benefits from the better fusion which has a global common objective and multi-scale concept.
$\textit{MSDTrans}$ improves the performance from $\textit{GAME0}$ (11.62) to $\textit{GAME0}$ (11.17).
It indicates the supplementary effect of a modality on the other modality.
Last, the whole model achieves a best $\textit{GAME0}$ (10.90), which shows the effectiveness of both modules.
However, we also find that $\textit{RMSE}$ value in the second line achieves the best result. It suggests our future work to improve the model.

\begin{table*}[!htp]
\caption{Ablation study about modules. The best result is in bold.}
\centering
\resizebox{\linewidth}{!}
{
\begin{tabular}{c|ccc|c|c|c|c|c}
    \hline\toprule
   \multirow{2}{*}{\centering Variant} & \multicolumn{3}{c|}{\centering Candidate} & \multirow{2}{*}{\centering GAME(0)} & \multirow{2}{*}{\centering GAME(1)} & \multirow{2}{*}{\centering GAME(2)} & \multirow{2}{*}{\centering GAME(3)} & \multirow{2}{*}{\centering RMSE}\\
     & Baseline  & MSTTrans &MSDTrans & & & & &\\
    \hline
    No.1  &$\checkmark$ & &
     & 11.62 & 16.25 & 20.38 & 27.17 & 19.88 \\
    No.2  &$\checkmark$ &$\checkmark$ &
     & 10.91 & 15.26 & 19.88 & 26.99 &\textbf{18.32} \\
    No.3  &$\checkmark$ &&$\checkmark$
     & 11.22 & 15.20 & 19.42 & 26.30&19.75 \\
     No.4  &$\checkmark$ & $\checkmark$&$\checkmark$
     & \textbf{10.90} & \textbf{14.81} & \textbf{19.02} &\textbf{26.14} &18.79\\
    \bottomrule
    \hline
\end{tabular}
}
\label{tab:ModuleAblation}
\end{table*}

\subsubsection{Effectiveness analysis of the count-guided multi-modal fusion design}
To verify  our contributions, we conduct the ablation studies about the count-guided multi-modal fusion design.
There are two essential design conceptions in the module.
One is the guidance of the learnable count token. The other is multi-scale strategy.
Table \ref{tab:DesignAblation} show the result.
At first, we show our result in the first line.
Then, we remove the learnable count token from the whole model.
Finally, we replace the multi-scale token transformer with vanilla multi-head self-attention.
By the observation, we find that the performance declines obviously when removing the count token. It just verifies the effectiveness of the count token.
Furthermore, multi-scale concept is also effective because the performance is worse when replacing our proposed multi-scale token transformer with multi-head self-attention. Compared with both, multi-scale concept plays a more important role than the learnable count token. It also verifies our most important contribution which introduces a token level multi-scale transformer.
\begin{table*}[!htp]
\caption{Ablation study about count guidance and multi-scale concept  in count-guided multi-modal fusion module. The best result is in bold.  ``Ours/count" represents our model removing the learnable count token. ``Ours/multi-scale" represents our model with vanilla multi-head self-attention instead of the multi-scale token transformer.}
\centering
\resizebox{\linewidth}{!}
{
\begin{tabular}{c|ccc|c|c|c|c|c}
    \hline\toprule
   \multirow{2}{*}{\centering Variant} & \multicolumn{3}{c|}{\centering Candidate} & \multirow{2}{*}{\centering GAME(0)} & \multirow{2}{*}{\centering GAME(1)} & \multirow{2}{*}{\centering GAME(2)} & \multirow{2}{*}{\centering GAME(3)} & \multirow{2}{*}{\centering RMSE}\\
   &  Ours  & Ours/count &Ours/multi-scale     & & & & &\\
    \hline
    No.1  &$\checkmark$ & &
     & \textbf{10.90} & \textbf{14.81} & \textbf{19.02} &\textbf{26.14} &\textbf{18.79} \\
    No.2  & &$\checkmark$ &
     & 11.82 & 15.91 & 20.10 & 27.13 &20.54 \\

     No.3  & & &$\checkmark$
      & 11.82 & 16.39 & 20.89 & 28.37 & 21.73\\
    \bottomrule
    \hline
\end{tabular}
}
\label{tab:DesignAblation}
\end{table*}

\section{Conclusions}
In the paper, we propose an RGB-T multi-modal crowd counting method
based on Transformer. Two-modal features are fused under the guidance of a learnable count token. Then crowd density map is predicted  by a modality and guided by the other modality. To solve the large-scale variation problem, a multi-scale token transformer is proposed to diversify the receptive fields.
The experimental results demonstrate a significant improvement over existing RGB-T crowd counting methods and verify the effectiveness of all the designs.
\section{Acknowledgment}
This work is supported by Natural Science Foundation of Anhui Province (1908085MF182) and Science Research Project for Graduate Student of Anhui Provincial Education Department (YJS20210047).
\bibliography{TransRGBTbib}

\begin{thebibliography}{50}
\providecommand{\natexlab}[1]{#1}
\providecommand{\url}[1]{\texttt{#1}}
\expandafter\ifx\csname urlstyle\endcsname\relax
  \providecommand{\doi}[1]{doi: #1}\else
  \providecommand{\doi}{doi: \begingroup \urlstyle{rm}\Url}\fi

\bibitem[Babu~Sam et~al.(2017)Babu~Sam, Surya, and
  Venkatesh~Babu]{babu2017switching}
Deepak Babu~Sam, Shiv Surya, and R~Venkatesh~Babu.
\newblock {Switching Convolutional Neural Network for Crowd Counting}.
\newblock In \emph{Proceedings of the IEEE Conference on Computer Vision and
  Pattern Recognition}, pages 5744--5752, 2017.

\bibitem[Bai et~al.(2020)Bai, He, Qiao, Hu, Wu, and Yan]{bai2020adaptive}
Shuai Bai, Zhiqun He, Yu~Qiao, Hanzhe Hu, Wei Wu, and Junjie Yan.
\newblock {Adaptive Dilated Network with Self-Correction Supervision for
  Counting}.
\newblock In \emph{Proceedings of the IEEE/CVF Conference on Computer Vision
  and Pattern Recognition}, pages 4594--4603, 2020.

\bibitem[Chen et~al.(2022)Chen, Wang, Su, and Wang]{chen2022ssr}
Jiwei Chen, Kewei Wang, Wen Su, and Zengfu Wang.
\newblock {SSR-HEF: Crowd Counting with Multi-Scale Semantic Refining and Hard
  Example Focusing}.
\newblock \emph{IEEE Transactions on Industrial Informatics}, pages 6547--6557,
  2022.

\bibitem[Chen et~al.(2017)Chen, Papandreou, Kokkinos, Murphy, and
  Yuille]{chen2017deeplab}
Liang-Chieh Chen, George Papandreou, Iasonas Kokkinos, Kevin Murphy, and Alan~L
  Yuille.
\newblock Deeplab: Semantic image segmentation with deep convolutional nets,
  atrous convolution, and fully connected crfs.
\newblock \emph{IEEE transactions on pattern analysis and machine
  intelligence}, 40\penalty0 (4):\penalty0 834--848, 2017.

\bibitem[Chen and Lu(2022)]{chen2022reinforcing}
Xiaoshuang Chen and Hongtao Lu.
\newblock {Reinforcing Local Feature Representation for Weakly-Supervised Dense
  Crowd Counting}.
\newblock \emph{arXiv preprint arXiv:2202.10681}, 2022.

\bibitem[Dai et~al.(2021)Dai, Liu, Ma, Zhang, and Zhao]{dai2021dense}
Feng Dai, Hao Liu, Yike Ma, Xi~Zhang, and Qiang Zhao.
\newblock {Dense Scale Network for Crowd Counting}.
\newblock In \emph{Proceedings of the 2021 International Conference on
  Multimedia Retrieval}, pages 64--72, 2021.

\bibitem[Fan et~al.(2022)Fan, Zhang, Zhang, Lu, Zhang, and Wang]{fan2022survey}
Zizhu Fan, Hong Zhang, Zheng Zhang, Guangming Lu, Yudong Zhang, and Yaowei
  Wang.
\newblock {A Survey of Crowd Counting and Density Estimation based on
  Convolutional Neural Network}.
\newblock \emph{Neurocomputing}, 472:\penalty0 224--251, 2022.

\bibitem[Gao et~al.(2020)Gao, Gao, Liu, Wang, and Wang]{gao2020cnn}
Guangshuai Gao, Junyu Gao, Qingjie Liu, Qi~Wang, and Yunhong Wang.
\newblock {CNN-Based Density Estimation and Crowd Counting: A Survey}.
\newblock \emph{arXiv preprint arXiv:2003.12783}, 2020.

\bibitem[Gao et~al.(2019)Gao, Wang, and Li]{gao2019pcc}
Junyu Gao, Qi~Wang, and Xuelong Li.
\newblock Pcc net: Perspective crowd counting via spatial convolutional
  network.
\newblock \emph{IEEE Transactions on Circuits and Systems for Video
  Technology}, 30\penalty0 (10):\penalty0 3486--3498, 2019.

\bibitem[Gao et~al.(2022)Gao, Gong, and Li]{gao2021congested}
Junyu Gao, Maoguo Gong, and Xuelong Li.
\newblock Congested crowd instance localization with dilated convolutional swin
  transformer.
\newblock \emph{Neurocomputing}, pages 94--103, 2022.

\bibitem[Guerrero-G{\'o}mez-Olmedo et~al.(2015)Guerrero-G{\'o}mez-Olmedo,
  Torre-Jim{\'e}nez, L{\'o}pez-Sastre, Maldonado-Basc{\'o}n, and
  Onoro-Rubio]{guerrero2015extremely}
Ricardo Guerrero-G{\'o}mez-Olmedo, Beatriz Torre-Jim{\'e}nez, Roberto
  L{\'o}pez-Sastre, Saturnino Maldonado-Basc{\'o}n, and Daniel Onoro-Rubio.
\newblock {Extremely Overlapping Vehicle Counting}.
\newblock In \emph{Iberian Conference on Pattern Recognition and Image
  Analysis}, pages 423--431. Springer, 2015.

\bibitem[Hoai and Zisserman(2014)]{hoai2014talking}
Minh Hoai and Andrew Zisserman.
\newblock {Talking Heads: Detecting Humans and Recognizing Their Interactions}.
\newblock In \emph{Proceedings of the IEEE Conference on Computer Vision and
  Pattern Recognition}, pages 875--882, 2014.

\bibitem[Idrees et~al.(2015)Idrees, Soomro, and Shah]{idrees2015detecting}
Haroon Idrees, Khurram Soomro, and Mubarak Shah.
\newblock {Detecting Humans in Dense Crowds Using Locally-Consistent Scale
  Prior and Global Occlusion Reasoning}.
\newblock \emph{IEEE transactions on pattern analysis and machine
  intelligence}, 37\penalty0 (10):\penalty0 1986--1998, 2015.

\bibitem[Jiang et~al.(2020)Jiang, Zhang, Zhang, Lv, Zhou, Pang, Xu, and
  Xu]{jiang2020density}
Xiaoheng Jiang, Li~Zhang, Tianzhu Zhang, Pei Lv, Bing Zhou, Yanwei Pang,
  Mingliang Xu, and Changsheng Xu.
\newblock {Density-Aware Multi-Task Learning for Crowd Counting}.
\newblock \emph{IEEE Transactions on Multimedia}, 23:\penalty0 443--453, 2020.

\bibitem[Li et~al.(2018)Li, Zhang, and Chen]{li2018csrnet}
Yuhong Li, Xiaofan Zhang, and Deming Chen.
\newblock {CSRNet: Dilated Convolutional Neural Networks for Understanding the
  Highly Congested Scenes}.
\newblock In \emph{Proceedings of the IEEE conference on computer vision and
  pattern recognition}, pages 1091--1100, 2018.

\bibitem[Lian et~al.(2021)Lian, Chen, Li, Luo, and Gao]{lian2021locating}
Dongze Lian, Xianing Chen, Jing Li, Weixin Luo, and Shenghua Gao.
\newblock Locating and counting heads in crowds with a depth prior.
\newblock \emph{IEEE Transactions on Pattern Analysis and Machine
  Intelligence}, pages 1--17, 2021.

\bibitem[Liang et~al.(2022)Liang, Chen, Xu, Zhou, and Bai]{liang2022transcrowd}
Dingkang Liang, Xiwu Chen, Wei Xu, Yu~Zhou, and Xiang Bai.
\newblock {TransCrowd: Weakly-Supervised Crowd Counting with Transformers}.
\newblock \emph{Science China Information Sciences}, 65\penalty0 (6):\penalty0
  1--14, 2022.

\bibitem[Lin et~al.(2022)Lin, Ma, Ji, Wang, and Hong]{lin2022boosting}
Hui Lin, Zhiheng Ma, Rongrong Ji, Yaowei Wang, and Xiaopeng Hong.
\newblock {Boosting Crowd Counting via Multifaceted Attention}.
\newblock In \emph{Proceedings of the IEEE/CVF Conference on Computer Vision
  and Pattern Recognition}, pages 19628--19637, 2022.

\bibitem[Liu et~al.(2020)Liu, Zhen, Li, Zhan, He, Du, and Lin]{liu2020dynamic}
Lingbo Liu, Jiajie Zhen, Guanbin Li, Geng Zhan, Zhaocheng He, Bowen Du, and
  Liang Lin.
\newblock {Dynamic Spatial-Temporal Representation Learning for Traffic Flow
  Prediction}.
\newblock \emph{IEEE Transactions on Intelligent Transportation Systems},
  22\penalty0 (11):\penalty0 7169--7183, 2020.

\bibitem[Liu et~al.(2021)Liu, Chen, Wu, Li, Li, and Lin]{liu2021cross}
Lingbo Liu, Jiaqi Chen, Hefeng Wu, Guanbin Li, Chenglong Li, and Liang Lin.
\newblock {Cross-Modal Collaborative Representation Learning and a Large-Scale
  RGBT Benchmark for Crowd Counting}.
\newblock In \emph{Proceedings of the IEEE/CVF Conference on Computer Vision
  and Pattern Recognition}, pages 4823--4833, 2021.

\bibitem[Liu et~al.(2019)Liu, Salzmann, and Fua]{liu2019context}
Weizhe Liu, Mathieu Salzmann, and Pascal Fua.
\newblock {Context-aware crowd counting}.
\newblock In \emph{Proceedings of the IEEE/CVF conference on computer vision
  and pattern recognition}, pages 5099--5108, 2019.

\bibitem[Ma et~al.(2019)Ma, Wei, Hong, and Gong]{ma2019bayesian}
Zhiheng Ma, Xing Wei, Xiaopeng Hong, and Yihong Gong.
\newblock {Bayesian Loss for Crowd Count Estimation with Point Supervision}.
\newblock In \emph{Proceedings of the IEEE/CVF International Conference on
  Computer Vision}, pages 6142--6151, 2019.

\bibitem[Rong and Li(2021)]{rong2021coarse}
Liangzi Rong and Chunping Li.
\newblock {Coarse- and Fine-Grained Attention Network with Background-Aware
  Loss for Crowd Density Map Estimation}.
\newblock In \emph{Proceedings of the IEEE/CVF Winter Conference on
  Applications of Computer Vision}, pages 3675--3684, 2021.

\bibitem[Sajid et~al.(2021)Sajid, Chen, Sajid, Kim, and Wang]{sajid2021audio}
Usman Sajid, Xiangyu Chen, Hasan Sajid, Taejoon Kim, and Guanghui Wang.
\newblock {Audio-visual transformer based crowd counting}.
\newblock In \emph{Proceedings of the IEEE/CVF International Conference on
  Computer Vision}, pages 2249--2259, 2021.

\bibitem[Shi et~al.(2019)Shi, Yang, Xu, and Chen]{shi2019revisiting}
Miaojing Shi, Zhaohui Yang, Chao Xu, and Qijun Chen.
\newblock {Revisiting Perspective Information for Efficient Crowd Counting}.
\newblock In \emph{Proceedings of the IEEE/CVF Conference on Computer Vision
  and Pattern Recognition}, pages 7279--7288, 2019.

\bibitem[Song et~al.(2021)Song, Wang, Jiang, Wang, Tai, Wang, Li, Huang, and
  Wu]{song2021rethinking}
Qingyu Song, Changan Wang, Zhengkai Jiang, Yabiao Wang, Ying Tai, Chengjie
  Wang, Jilin Li, Feiyue Huang, and Yang Wu.
\newblock {Rethinking Counting and Localization in Crowds: A Purely Point-Based
  Framework}.
\newblock In \emph{Proceedings of the IEEE/CVF International Conference on
  Computer Vision}, pages 3365--3374, 2021.

\bibitem[Stewart et~al.(2016)Stewart, Andriluka, and Ng]{stewart2016end}
Russell Stewart, Mykhaylo Andriluka, and Andrew~Y Ng.
\newblock {End-to-End People Detection in Crowded Scenes}.
\newblock In \emph{Proceedings of the IEEE conference on computer vision and
  pattern recognition}, pages 2325--2333, 2016.

\bibitem[Sun et~al.(2021)Sun, Liu, Probst, Paudel, Popovic, and
  Van~Gool]{sun2021boosting}
Guolei Sun, Yun Liu, Thomas Probst, Danda~Pani Paudel, Nikola Popovic, and Luc
  Van~Gool.
\newblock {Boosting Crowd Counting with Transformers}.
\newblock \emph{arXiv preprint arXiv:2105.10926}, 2021.

\bibitem[Tang et~al.(2022)Tang, Wang, and Chau]{tang2022tafnet}
Haihan Tang, Yi~Wang, and Lap-Pui Chau.
\newblock {TAFNet: A Three-Stream Adaptive Fusion Network for RGB-T Crowd
  Counting}.
\newblock \emph{arXiv preprint arXiv:2202.08517}, 2022.

\bibitem[Tian et~al.(2021)Tian, Chu, and Wang]{tian2021cctrans}
Ye~Tian, Xiangxiang Chu, and Hongpeng Wang.
\newblock {CCTrans: Simplifying and Improving Crowd Counting with Transformer}.
\newblock \emph{arXiv preprint arXiv:2109.14483}, 2021.

\bibitem[Usman and Albesher(2021)]{usman2021abnormal}
Imran Usman and Abdulaziz~A Albesher.
\newblock {Abnormal Crowd Behavior Detection Using Heuristic Search and Motion
  Awareness}.
\newblock \emph{International Journal of Computer Science \& Network Security},
  21\penalty0 (4):\penalty0 131--139, 2021.

\bibitem[Velavan and Meyer(2020)]{velavan2020covid}
Thirumalaisamy~P Velavan and Christian~G Meyer.
\newblock {The COVID-19 Epidemic}.
\newblock \emph{Tropical Medicine \& International Health}, 25\penalty0
  (3):\penalty0 278, 2020.

\bibitem[Wang et~al.(2020)Wang, Liu, Samaras, and Nguyen]{wang2020distribution}
Boyu Wang, Huidong Liu, Dimitris Samaras, and Minh~Hoai Nguyen.
\newblock {Distribution Matching for Crowd Counting}.
\newblock \emph{Advances in Neural Information Processing Systems},
  33:\penalty0 1595--1607, 2020.

\bibitem[Wang et~al.(2022{\natexlab{a}})Wang, Liu, Long, Sang, Xia, and
  Sang]{wang2022joint}
Fusen Wang, Kai Liu, Fei Long, Nong Sang, Xiaofeng Xia, and Jun Sang.
\newblock {Joint CNN and Transformer Network via Weakly Supervised Learning for
  Efficient Crowd Counting}.
\newblock \emph{arXiv preprint arXiv:2203.06388}, 2022{\natexlab{a}}.

\bibitem[Wang et~al.(2022{\natexlab{b}})Wang, Sang, Wu, Liu, and
  Sang]{wang2022hybrid}
Fusen Wang, Jun Sang, Zhongyuan Wu, Qi~Liu, and Nong Sang.
\newblock {Hybrid Attention Network Based on Progressive Embedding
  Scale-Context for Crowd Counting}.
\newblock \emph{Information Sciences}, 591:\penalty0 306--318,
  2022{\natexlab{b}}.

\bibitem[Wang et~al.(2022{\natexlab{c}})Wang, Cai, Han, Zhou, and
  Gong]{wang2022stnet}
Mingjie Wang, Hao Cai, Xianfeng Han, Jun Zhou, and Minglun Gong.
\newblock {STNet: Scale Tree Network with Multi-level Auxiliator for Crowd
  Counting}.
\newblock \emph{IEEE Transactions on Multimedia}, pages 1--11,
  2022{\natexlab{c}}.

\bibitem[Wang et~al.(2022{\natexlab{d}})Wang, Zhou, Cai, and
  Gong]{wang2022crowdmlp}
Mingjie Wang, Jun Zhou, Hao Cai, and Minglun Gong.
\newblock {CrowdMLP: Weakly-Supervised Crowd Counting via Multi-Granularity
  MLP}.
\newblock \emph{arXiv preprint arXiv:2203.08219}, 2022{\natexlab{d}}.

\bibitem[Wang and Breckon(2022)]{wang2022crowd}
Qian Wang and Toby~P Breckon.
\newblock {Crowd Counting via Segmentation Guided Attention Networks and
  Curriculum Loss}.
\newblock \emph{IEEE Transactions on Intelligent Transportation Systems}, pages
  15233--15243, 2022.

\bibitem[Wang et~al.(2021)Wang, Xie, Li, Fan, Song, Liang, Lu, Luo, and
  Shao]{wang2021pyramid}
Wenhai Wang, Enze Xie, Xiang Li, Deng-Ping Fan, Kaitao Song, Ding Liang, Tong
  Lu, Ping Luo, and Ling Shao.
\newblock {Pyramid Vision Transformer: A Versatile Backbone for Dense
  Prediction without Convolutions}.
\newblock In \emph{Proceedings of the IEEE/CVF International Conference on
  Computer Vision}, pages 568--578, 2021.

\bibitem[Wei et~al.(2021)Wei, Kang, Yang, Qiu, Shi, Tan, and
  Gong]{wei2021scene}
Xing Wei, Yuanrui Kang, Jihao Yang, Yunfeng Qiu, Dahu Shi, Wenming Tan, and
  Yihong Gong.
\newblock {Scene-Adaptive Attention Network for Crowd Counting}.
\newblock \emph{arXiv preprint arXiv:2112.15509}, 2021.

\bibitem[Wu et~al.(2022)Wu, Liu, Zhang, Mao, Lin, and Li]{wu2022multimodal}
Zhengtao Wu, Lingbo Liu, Yang Zhang, Mingzhi Mao, Liang Lin, and Guanbin Li.
\newblock {Multimodal Crowd Counting with Mutual Attention Transformers}.
\newblock In \emph{2022 IEEE International Conference on Multimedia and Expo
  (ICME)}, pages 1--6. IEEE, 2022.

\bibitem[Yan et~al.(2019)Yan, Yuan, Zuo, Tan, Wang, Wen, and
  Ding]{yan2019perspective}
Zhaoyi Yan, Yuchen Yuan, Wangmeng Zuo, Xiao Tan, Yezhen Wang, Shilei Wen, and
  Errui Ding.
\newblock {Perspective-Guided Convolution Networks for Crowd Counting}.
\newblock In \emph{Proceedings of the IEEE/CVF International Conference on
  Computer Vision}, pages 952--961, 2019.

\bibitem[Yang et~al.(2022)Yang, Guo, and Ren]{yangcrowdformer}
Shangpeng Yang, Weiyu Guo, and Yuheng Ren.
\newblock {CrowdFormer: An Overlap Patching Vision Transformer for Top-Down
  Crowd Counting}.
\newblock In \emph{Proceedings of the Thirty-First International Joint
  Conference on Artificial Intelligence (IJCAI-22)}, pages 1545--1551, 2022.

\bibitem[Yang et~al.(2020{\natexlab{a}})Yang, Li, Du, Huang, and
  Sebe]{yang2020embedding}
Yifan Yang, Guorong Li, Dawei Du, Qingming Huang, and Nicu Sebe.
\newblock {Embedding Perspective Analysis into Multi-Column Convolutional
  Neural Network for Crowd Counting}.
\newblock \emph{IEEE Transactions on Image Processing}, 30:\penalty0
  1395--1407, 2020{\natexlab{a}}.

\bibitem[Yang et~al.(2020{\natexlab{b}})Yang, Li, Wu, Su, Huang, and
  Sebe]{yang2020reverse}
Yifan Yang, Guorong Li, Zhe Wu, Li~Su, Qingming Huang, and Nicu Sebe.
\newblock {Reverse Perspective Network for Perspective-Aware Object Counting}.
\newblock In \emph{Proceedings of the IEEE/CVF Conference on Computer Vision
  and Pattern Recognition}, pages 4374--4383, 2020{\natexlab{b}}.

\bibitem[Yu et~al.(2022)Yu, Liang, Lin, Wan, Wang, and Dai]{yu2022frequency}
Xiaoyuan Yu, Yanyan Liang, Xuxin Lin, Jun Wan, Tian Wang, and Hong-Ning Dai.
\newblock {Frequency Feature Pyramid Network With Global-Local Consistency Loss
  for Crowd-and-Vehicle Counting in Congested Scenes}.
\newblock \emph{IEEE Transactions on Intelligent Transportation Systems}, pages
  9654--9664, 2022.

\bibitem[Yuan et~al.(2020)Yuan, Qiu, Liu, Wu, Chen, Chen, and
  Lin]{yuan2020crowd}
Lixian Yuan, Zhilin Qiu, Lingbo Liu, Hefeng Wu, Tianshui Chen, Pei Chen, and
  Liang Lin.
\newblock {Crowd Counting via Scale-Communicative Aggregation Networks}.
\newblock \emph{Neurocomputing}, 409:\penalty0 420--430, 2020.

\bibitem[Zhang et~al.(2016)Zhang, Zhou, Chen, Gao, and Ma]{zhang2016single}
Yingying Zhang, Desen Zhou, Siqin Chen, Shenghua Gao, and Yi~Ma.
\newblock {Single-Image Crowd Counting via Multi-Column Convolutional Neural
  Network}.
\newblock In \emph{Proceedings of the IEEE Conference on Computer Vision and
  Pattern Recognition}, pages 589--597. CVPR, 2016.

\bibitem[Zhou et~al.(2022)Zhou, Pan, Lei, Ye, and Yu]{zhou2022defnet}
Wujie Zhou, Yi~Pan, Jingsheng Lei, Lv~Ye, and Lu~Yu.
\newblock {DEFNet: Dual-Branch Enhanced Feature Fusion Network for RGB-T Crowd
  Counting}.
\newblock \emph{IEEE Transactions on Intelligent Transportation Systems}, pages
  1--10, 2022.

\bibitem[Zhu et~al.(2020)Zhu, Su, Lu, Li, Wang, and Dai]{zhu2020deformable}
Xizhou Zhu, Weijie Su, Lewei Lu, Bin Li, Xiaogang Wang, and Jifeng Dai.
\newblock {Deformable DETR: Deformable Transformers for End-to-End Object
  Detection}.
\newblock In \emph{International Conference on Learning Representations}, pages
  1--12, 2020.

\end{thebibliography}
\end{document}